\documentclass[11pt]{article}

\usepackage{acl}

\usepackage{times}
\usepackage{latexsym}

\usepackage[T1]{fontenc}
\usepackage[utf8]{inputenc}

\usepackage{microtype}

\usepackage{amsthm}
\usepackage{amssymb}
\usepackage{amsmath}
\usepackage{amsfonts}
\usepackage{xcolor}
\usepackage{multirow}
\usepackage{tabularx}
\usepackage{booktabs}
\usepackage{subcaption}
\usepackage{graphicx}
\usepackage{xspace}
\usepackage{color,colortbl}
\usepackage{url}
\usepackage{xspace}
\usepackage{dsfont}
\usepackage{arydshln}
\definecolor{Gray}{gray}{0.92}

\newcommand{\nlup}{{\textsc{nlu++}}\xspace}
\newcommand{\banking}{{\textsc{banking}}\xspace}
\newcommand{\hotels}{{\textsc{hotels}}\xspace}
\newcommand{\all}{{\textsc{all}}\xspace}

\newcommand{\sparagraph}[1]{\noindent\textbf{#1.}}
\newcommand{\rparagraph}[1]{\vspace{1.5mm}\noindent\textbf{#1.}}

\definecolor{nice-red}{HTML}{E41A1C}
\definecolor{nice-orange}{HTML}{FF7F00}
\definecolor{nice-yellow}{HTML}{FFC020}
\definecolor{nice-green}{HTML}{4DAF4A}
\definecolor{nice-blue}{HTML}{377EB8}
\definecolor{nice-purple}{HTML}{984EA3}

\newcolumntype{Y}{>{\centering\arraybackslash}X}

\usepackage{todonotes}
\makeatletter
\newcommand*\iftodonotes{\if@todonotes@disabled\expandafter\@secondoftwo\else\expandafter\@firstoftwo\fi}
\makeatother

\title{NLU++: A Multi-Label, Slot-Rich, Generalisable Dataset for Natural Language Understanding in Task-Oriented Dialogue}

\author{
 I{\~{n}}igo Casanueva\thanks{{ } Equal contribution.},
 Ivan Vuli\'{c},$^{*}$
 Georgios P. Spithourakis, \textmd{and}
 Pawe\l~Budzianowski \\
 PolyAI Limited \\
 London, United Kingdom \\
 \texttt{\{inigo,ivan,georgios,pawel\}@poly.ai}
}

\begin{document}
\maketitle
\begin{abstract}
We present \nlup, a novel dataset for natural language understanding (NLU) in task-oriented dialogue (ToD) systems, with the aim to provide a much more challenging evaluation environment for dialogue NLU models, up to date with the current application and industry requirements. \nlup is divided into two domains (\banking and \hotels) and brings several crucial improvements over current commonly used NLU datasets. \textbf{1)} \nlup provides fine-grained domain ontologies with a large set of challenging \textit{multi-intent} sentences, introducing and validating the idea of \textit{intent modules} that can be combined into complex intents that convey complex user goals, combined with finer-grained and thus more challenging slot sets. \textbf{2)} The ontology is divided into \textit{domain-specific} and \textit{generic} (i.e., domain-universal) intent modules that overlap across domains, promoting cross-domain reusability of annotated examples. \textbf{3)} The dataset design has been inspired by the problems observed in industrial ToD systems, and \textbf{4)} it has been collected, filtered and carefully annotated by dialogue NLU experts, yielding high-quality annotated data. Finally, we benchmark a series of current state-of-the-art NLU models on \nlup; the results demonstrate the challenging nature of the dataset, especially in low-data regimes, the validity of `intent modularisation', and call for further research on ToD NLU.
\end{abstract}

\section{Introduction}
\label{sec:intro}
Research on task-oriented dialogue (ToD) systems \cite{Levin:95, Young:2002} has become a key aspect in industry: e.g., ToD is used to automate telephone customer service tasks ranging from hospitality over healthcare to banking \cite{Raux:2003,young:10,ElAsri:2017sigdial}. Typical ToD systems still rely on a modular design: (i) \textit{the natural language understanding} (NLU) module maps user utterances into a domain-specific set of intent labels and values \cite{rastogi2019towards, heck2020trippy, dai2021preview}% Mrksic:17}
, followed by (ii) the \textit{policy} module, which makes decisions based on the information extracted by the NLU \cite{gasic:10, casanueva2017benchmarking,lubis2020lava,wang2020multi}

\begin{figure}[t]\label{fig:nlu-intents-and-values}
\centering
\includegraphics[width=0.47\textwidth]{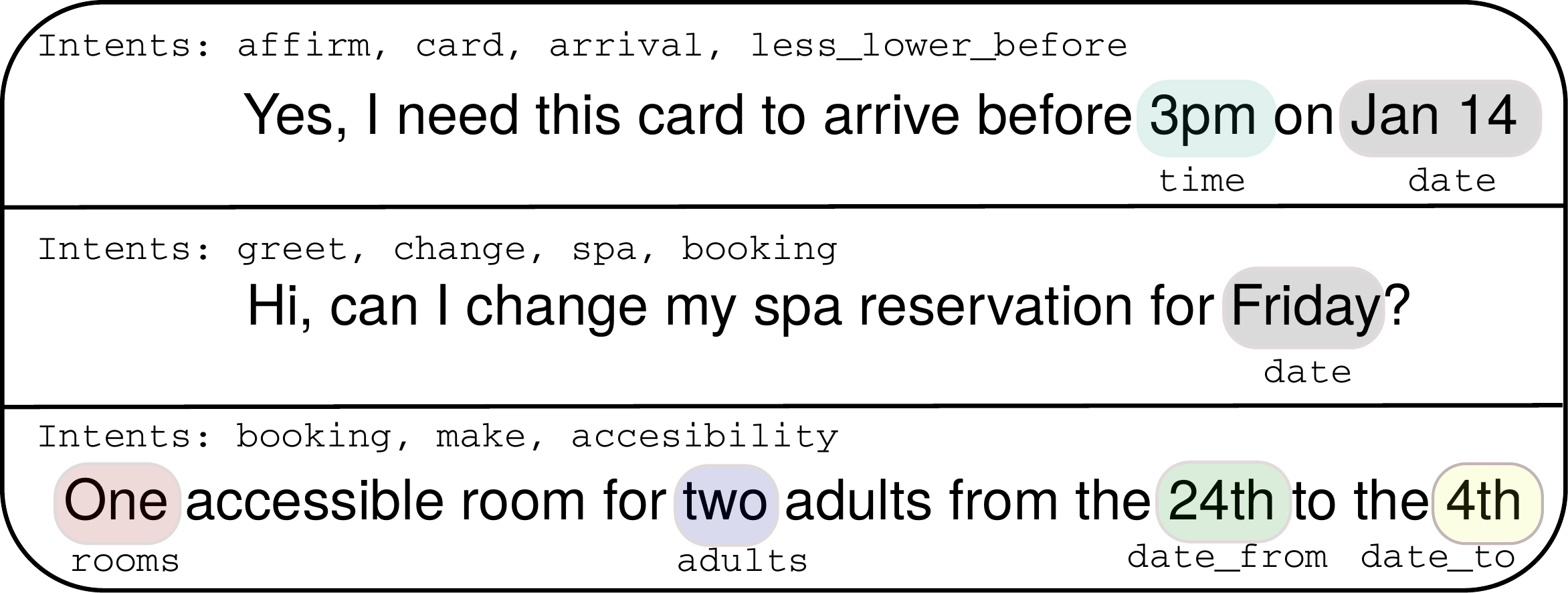}
%\vspace{-0.5mm}
\caption{Multi-intent examples from the two domains of the \nlup dataset: \banking (top) and \hotels (middle, bottom), illustrating the two core NLU subtasks of intent detection (ID) and slot labeling (SL) in ToD systems. The extracted information is structured into \textit{intents} and \textit{slots}, the latter having associated \textit{values}.}
%\paw{more sentences here - people love this. Also the figure proves nothing about the complexity of slots and we can show crazy time examples here (Inigo - could you provide some complicated ones?)}
\label{figure1}
%\vspace{-1.5mm}
\end{figure}

The NLU module is a critical part of any ToD system, as it must extract the \textit{relevant} information from the user's utterances. The information relevance is denoted by the structured \textit{dialogue domain ontology}, which enables the policy module to make decisions about next system actions. The domain ontology covers the information on 1) \textit{intents} and 2) \textit{slots}, see Figure~\ref{figure1}. The former is aimed at extracting general conversational ideas (i.e., the user's intents) and corresponds to the standard NLU task of \textit{intent detection (ID)}; the latter extracts specific \textit{slot values} and corresponds to the NLU task of  \textit{slot labeling (SL)} \cite{gupta2019simple}.\footnote{Slot labeling is also known under other names such as slot filling or value extraction.} 

\begin{table*}[!t]
\footnotesize
\def\arraystretch{0.83}
\centering
\begin{tabularx}{\linewidth}{l l X}
\toprule
\textbf{Example} & \textbf{Traditional Intent}  & \textbf{Intent Modules} \\
\midrule
\textit{I need to change my restaurant reservation} &change\_restaurant\_booking & change, restaurant, booking \\
\\
%\vspace{-1mm}
\textit{When is my booking for the spa?} &when\_spa\_booking & when, spa, booking \\
\\
%\vspace{-1mm}
\textit{TV is not showing any image} &tv\_not\_working  & tv, not\_working\\
\\
%\vspace{-1mm}
\textit{Why can't I cancel this standing order?} &why\_cancel\_standing\_order\_not\_working &why, cancel, standing\_order, not\_working \\
\bottomrule
\end{tabularx}
%\vspace{-1.5mm}
\caption{Comparison of "traditional" intent annotations vs \textit{intent module}-based multi-label annotations.}
\label{tab:intent-comparison}
%\vspace{-2mm}
\end{table*}

In order to make the policy operational \textit{and} tractable, NLU should extract only the minimal information required by the policy. Therefore, the ontologies differ for each domain of ToD application and are typically built from scratch for each domain. Consequently, this makes domain-relevant NLU data extremely expensive to collect and annotate, and prevents its reusability \cite{Budzianowski:2018emnlp}. Due to this, NLU research in recent years has heavily focused on very data-efficient models that can effectively operate in low-data regimes. Current state-of-the-art (SotA) NLU models leverage large pretrained language models (PLMs) \cite{Devlin:2018arxiv, Liu:2019roberta, Henderson:2020convert} and fine-tune them with small task-specific datasets \cite{Larson:2019emnlp, Casanueva:2020ws,Coucke:18}%, CoopeFarghly2020}.

%% \ivan{Do we have some reference to support this?}

%%\ivan{For Figure 1: use a multi-label examples (with 2 and 3 intents at least) from NLU++, not soem random examples from previous datasets... INIGO: I think this would be too cluttered and we already give NLU++ examples in section 3}

At the same time, the progress in creation of NLU datasets has not kept up with the impressive pace of NLU methodology development. However, designing domain ontologies and NLU datasets is also critical for steering further progress in NLU, both from methodology and application perspective. Put simply, current publicly available NLU datasets do not keep up to date with current industry/application requirements for many reasons. \textbf{1)} They are usually crowdsourced by untrained annotators (thus typically optimised for quantity rather than quality), yielding examples with low lexical diversity and prone to annotation errors. \textbf{2)} They typically assume one intent per example, and thus enable only much simpler single-label ID experiments; such setups are not realistic in more complex industry settings (see Figure~\ref{figure1} again) and lead to unnecessarily large intent sets. \textbf{3)} Their ontologies are tied to specific domains, making it difficult to reuse already available annotated data in other domains. \textbf{4)} The complexity of the defined tasks and ontologies is limited; the undesired artefact is that current NLU datasets might overestimate the NLU models' abilities, and are not able to separate models any more performance-wise.\footnote{For instance, for some standard and commonly used NLU datasets such as ATIS \cite{Hemphill:1990,Xu:2020emnlp} and SNIPS \cite{Coucke:18}, the results of SotA models are all in the region of 97-98 $F_1$, with new models getting statistically insignificant gains which might be due to overfitting to the test set or even some remaining annotation errors.}

 %% IV: I don't want to emphasise this 'joint NLU models' thing as we don't do any joint modeling here, and there are some joint NLU models out there; it's better to hide it.
%\textbf{4)} They usually focus on either intent classification or slot filling, not allowing the research in joint NLU models.

\iffalse
\begin{itemize}
  \item They are usually crowdsourced by untrained annotators, leading to examples with low lexical diversity and prone to annotation errors.
  \item They assume one intent per example, which is not realistic in complex industry settings and leads to unnecessarily large intent sets.
  \item Their ontologies are tied to specific domains, making it difficult to reuse the data in other systems.
  \item They usually focus on either intent classification or slot filling, not allowing the research in joint NLU models.
  \item The complexity of the tasks defined is limited, overestimating models' abilities and not being able to separate models any more\footnote{: For some datasets like \cite{Hemphill:1990,Coucke:18} results are in the regions of 97-98 F1, with new models getting statistically insignificant gains which might be due to overfitting to the test set.}.
\end{itemize}
\fi

In order to address all these gaps, we introduce \nlup, a novel NLU dataset which provides high-quality NLU data annotated by dialogue experts. \nlup provides \textit{multi-intent}, \textit{slot-rich} and \textit{semantically varied} NLU data, and is inspired by a number of NLU challenges which ToD systems typically face in production environments. Unlike previous ID datasets, examples are annotated with multiple labels, named \textit{intent modules}\footnote{Henceforth, whenever \textit{intents} are mentioned in the context of \nlup, we will be referring to \textit{intent modules}.} (see Table~\ref{tab:intent-comparison}), with some examples naturally obtaining even up to 6-7 labels. These labels can be seen as sub-intent annotations, where their combinations yield full intents equivalent to "traditional" intents (Table~\ref{tab:intent-comparison}). In addition, \nlup defines a rich set of slots which are combined with the multi-intent sentences. \nlup is divided into two domains (\banking and \hotels) where the two domain ontologies blend a set of \textit{domain-specific} intents and slots with a set of \textit{generic} (i.e., domain-universal) intents and slots. This design makes a crucial step towards generalisation and data reusability in NLU. 

Finally, we run a series of experiments on \nlup with current SotA ID and SL models, demonstrating the challenging nature of \nlup and ample room for future improvement, especially in low-data setups. Our benchmark comparisons also demonstrate strong performance and shed new light on the (ability of) recently emerging QA-based NLU models \cite{qanlu,Fuisz:2022arxiv}, and warrant further research on ToD NLU. The \nlup dataset is available at: \href{https://github.com/PolyAI-LDN/task-specific-datasets/tree/master/nlupp}{\texttt{github.com/PolyAI-LDN/}} \href{https://github.com/PolyAI-LDN/task-specific-datasets/tree/master/nlupp}{\texttt{task-specific-datasets}}.

%%under \textsc{CC BY 4.0} license.

\begin{table*}[t]
\centering
\def\arraystretch{0.99}
{\footnotesize
\begin{tabularx}{\linewidth}{l c cYc cYc}
\toprule
\rowcolor{Gray}
  {} & {} & \multicolumn{3}{c}{\bf \textsc{Intents}} & \multicolumn{3}{c}{\bf \textsc{Slots}} \\
  \cmidrule(lr){3-5} \cmidrule(lr){6-8}
\textbf{Domain} & \textbf{Number of examples} & \textbf{Total} & \textbf{Generic} & \textbf{Avg. per example} & \textbf{Total} & \textbf{Generic} & \textbf{Avg. per example} \\
%\textbf{Domain} & {}  & \textbf{Total} & \textbf{Generic} & {} & \textbf{Total} & \textbf{Generic} & \textbf{example} \\
\cmidrule(lr){3-5} \cmidrule(lr){6-8}
\banking & {2,071} & {48} & {26} & {2.25} & {13} & {10} & {0.46} \\
\hotels & {1,009} & {40} & {26} & {1.52} & {14} & {10} & {1.03} \\
 \all & {3,080} & {62} & {26} & {2.01} & {17} & {10} & {0.65} \\
\bottomrule
\end{tabularx}
}%
\vspace{-1.5mm}
\caption{Key statistics of the \nlup dataset.}
\label{tab:dataset-stats}
\vspace{-2mm}
\end{table*}

%in state-of-the art NLU models show the challenging nature of the dataset and warrants further work on NLU.

%% .\ivan{We should also mention some key dataset stats already in the intro. INIGO: wouldnt this be repeating ourselves? we already go deep on that on sect 3 and we are already short on space} 

\section{Background and Motivation}
\label{sec:rw}

\sparagraph{A Brief History of NLU Datasets}
As a core module of ToD systems, NLU has been researched since the early 1990s, when the Airline Travel Information System (ATIS) project was started \cite{Hemphill:1990}, consisting of spoken queries on flight-related information.\footnote{Remarkably, ATIS is still considered at present as one of the main go-to datasets in NLU reserach. This is also reflected in the fact that the recent most popular dataset for multilingual dialogue NLU was obtained by simply translating English ATIS to 8 more languages \cite[MultiATIS++]{Xu:2020emnlp}.} Over the next two decades, very few NLU resources were released.\footnote{We note that some Question Classification \cite{hovy2001use}, Paraphrasing \cite{dolan2005automatically} and Semantic Text Similarity\cite{Agirre:2012semeval} datasets could be seen as the seed of modern ID datasets, but were not initially built for that purpose.} 

%\ivan{The structure of section 2 could be -> Paragraph 1: A Brief History of NLU; P2: Current NLU Trends; P3 Current Gaps in NLU Datasets}

The lack of ToD NLU resources ended in 2013, with the beginning of the `dialogue state tracking (DST) era' \cite{Williams:13a, Henderson:14a, DSTC4}. Instead of just classifying each turn of the user, DST deals with keeping track of the user's goal over the entire dialogue history, i.e., all the previous user and system turns. Several datasets where released during the DST challenges, all of them comprising simple intent sets (usually tagged as \textit{dialogue acts}).

%Meanwhile, the machine learning community was being revolutionised by the recent advances on Deep Learning (DL) \cite{krizhevsky:2012, Mikolov:13, sutskever:14}. 

In order to adapt to the increasing data requirements of deep learning models, increasingly larger dialogue datasets have been released in recent years \cite{Budzianowski:2018emnlp, wei2018airdialogue, rastogi2019towards,peskov2019multi}. However, the design of ToD datasets comes with some profound differences to datasets for e.g. machine translation or speech recognition, which affect current ToD datasets. \textbf{1)} The domain-specific nature of ToD datasets made the data tied to its ontologies, not allowing data reusability across different domains. \textbf{2)} The domain-specific ontologies required a lot of expertise for annotation, therefore many annotation mistakes were made \cite{Eric:2019arxiv, zang2020multiwoz}. \textbf{3)} Collecting datasets of that size is unfeasible for development cycles in production, where new domains and models for them need to be very quickly developed and deployed.

%\ivan{This is a nice paragraph (that needs language revision), but it's more "Motivation" or "Current NLU Trends" than "Related Work" INIGO: duno where to put it though, i think that fits well in the story even if its not technically related work. maybe call the section "background and motivation"?}

\vspace{1.5mm}
\noindent \textbf{Current NLU Trends}, inspired by such production requirements, thus
deviate from previous DST-oriented NLU research in two main aspects. First, the models went back to focusing on single-turn utterances, which \textbf{1)} simplifies the NLU design and \textbf{2)} renders the NLU tasks more tractable.\footnote{While DST is theoretically more accurate, it requires amounts of data that grow exponentially with the number of turns; moreover, rule-based trackers have proven to be on par with the learned/statistical ones and require no data \cite{Wang:13}.} %, Henderson:14a, Mrksic:17a}.} 
The requirement of fast development cycles also instigated more research on NLU (i.e., ID and SL tasks) in low-data scenarios. This way, systems can be developed and maintained faster by reducing the data collection and annotation effort. In addition, the NLU focus shifted from ontologies with only a handful of simple intents and slots \cite{Coucke:18} to complex ontologies with much larger intent sets \cite[\textit{inter alia}]{Larson:2019emnlp,Liu:2019iwsds,Casanueva:2020ws}.

Inspired by these NLU datasets and empowered by transfer learning with PLMs and sentence encoders \cite{Devlin:2018arxiv, Liu:2019arxiv, Henderson:2020convert}, there have been great improvements in single-turn NLU systems recently, especially in low-data scenarios \cite{CoopeFarghly2020, mehri2020example, wu-etal-2020-slotrefine, wu-etal-2020-tod, krone-etal-2020-learning, Henderson:2021convex, qanlu, dopierre-etal-2021-protaugment,zhang-etal-2021-effectiveness-pre,zhang-etal-2021-shot}. 

\begin{table*}[!t]
\footnotesize
\def\arraystretch{0.99}
\centering
\begin{tabularx}{\linewidth}{l l X}
\toprule
\textbf{Example} & \textbf{Intents}  & \textbf{Domain} \\
\midrule
\textit{I want to change my room reservation} &change, booking, room &\hotels \\
\textit{I want to cancel a booking} &cancel, booking &\hotels \\
\textit{Why can't I amend my restaurant booking?} &why, change, restaurant, booking, not\_working &\hotels \\
\hdashline
\textit{I am trying to make a transfer but it doesn't let me} &make, transfer\_payment, not\_working &\banking \\
\textit{I need to increase my overdraft} &change, overdraft, higher &\banking \\
\textit{Please close my savings account} &cancel, account, savings &\banking \\
\textit{The savings one} &savings &\banking \\
\hdashline
\textit{Make it higher} &change, higher & \textsc{general} \\
\textit{Cancel it} &cancel & \textsc{general} \\
\textit{Don't cancel it} &deny, cancel & \textsc{general} \\
\bottomrule
\end{tabularx}
%\vspace{-1.5mm}
\caption{\nlup examples showing the combinatorial expressiveness of intent modules in the multi-intent setting.}
\label{tab:multi-intent-examples} 
%\vspace{-2mm}
\end{table*}

\rparagraph{Current Gaps in NLU Datasets}
However, existing NLU datasets are still not up to the current industry requirements. \textbf{1)} They use crowdworkers for data collection and annotation, often through simple rephrasings; they thus suffer from low lexical diversity and annotation errors \cite{larson2019outlier}. \textbf{2)} ID datasets always assume a single intent per sentence, which does not support modern production requirements. \textbf{3)} The ontologies of these datasets are very domain-specific (i.e., they thus do not allow data reusability) and narrow (i.e., they tend to overestimate abilities of the current SotA NLU models). \textbf{4)} Current NLU datasets do not combine a large set of fine-grained intents (again, with multi-intent examples) and a large set of fine-grained slots, which prevents proper and more insightful evaluations of joint NLU models \cite{chen2019bert, gangadharaiah-narayanaswamy-2019-joint}.

We note that there has been some work on multi-label ID on ATIS, MultiWOZ and DSTC4 as multi-intent datasets; however, their multi-label examples remain very limited, simple, and span a small number of intents \cite{gangadharaiah-narayanaswamy-2019-joint}. Further, synthetic multi-intent datasets have been created by concatenating single-intent sentences, but such datasets also do not capture the complexity of true and natural multi-intent sentences \cite{qin2020agif}.

%ivan{INIGO: it would be good to have other papers that use this datasets cited so we dont look like idiots that only cithe their own work, but tbh im so out of the lloop that i dont know any other}

\section{\nlup Dataset}
\label{sec:nlupp}
The \nlup dataset has been designed with the aim of addressing some of the major shortcomings of the current NLU datasets. In what follows, we describe the main improvements and new evaluation opportunities offered by \nlup.

\subsection{Ontology}
\nlup comprises two domains: \banking and \hotels. The former represents a banking services task (e.g., making transfers, depositing cheques, reporting lost cards, requesting mortgage information) and the latter is a hotel `bell desk' reception task (e.g., booking rooms, asking about pools or gyms, requesting room service). Both domains combine a large set of intents with a rich set of slots, with the ontologies inspired by requirements in production. A large number of intents and slots is shared between the two domains, in an attempt to increase data reusability/transferability. Table \ref{tab:dataset-stats} provides the main statistics of the \nlup dataset, while the full ontology is presented in Appendix~\ref{sec:appendix1}.

%%  with \textsc{BANKING} leaning towards an intent-rich domain and \textsc{HOTELS} towards a slot-rich domain. 

\subsection{Multi-Intent Examples}

One of the main contributions of this work is the novel design of the intent space, defined in a highly modular manner that natively supports intent recombinations and multi-intent annotations\footnote{\citet{Zhang2020multipoint} proposed a similar way of annotating existing intent detection datasets, showing performance improvements. However, this approach forced categorising the sub-intents in four predefined factors.}. For instance, Table \ref{tab:multi-intent-examples} shows several multi-intent examples based on the intent sets (termed \textit{intent modules}) from Table~\ref{tab:intents-ontology} in Appendix~\ref{sec:appendix1}. 

This design brings several benefits. \textbf{1)} The modular nature of the ontology allows for expressing a much more complex set of ideas through different combinations of intent modules (see Table~\ref{tab:multi-intent-examples}), while reducing the overall size of the intent set compared to previous ID datasets\footnote{Similar to how sub-word tokenization reduced the size of language model vocabularies while covering a larger set of words \cite{vaswani2018}} (see Table~\ref{tab:intent-comparison} and Table~\ref{tab:dataset-comparison}). \textbf{2)} It allows for the definition of \textit{partial} intents (e.g., \textit{``The savings one''}). This is crucial in multi-turn interactions, where the user often has to answer disambiguation questions (e.g., \textit{``Which account would you like to close?''}). \textbf{3)} The modular approach allows the models to generalise to unseen combinations of intent modules. For instance, if (i) examples with the intents \textit{change} and \textit{booking}, and (ii) examples with the intents \textit{cancel} and \textit{account} exist in the training data, (iii) an unseen example with the intents \textit{cancel} and \textit{booking} could be properly predicted, as all the single intents/modules have already been seen by the ID model\footnote{Note that in single-label ID setups, all possible intent module combinations (i.e. "traditional" intents) must be covered \cite{bikwok,Hou:2020mlabel}, which leads to unnecessarily large intent sets and larger data requirements.}. \textbf{4)} The design also allows us to distinguish between \textit{domain-specific} versus \textit{generic} intent modules. For example, the module \textit{overdraft} is clearly related to \banking, but the module \textit{change} is much more generic, likely to occur in several different domains.

Finally, the modular design also allows us to study semantic variation of intent modules. Some intents (e.g., especially the domain-specific ones) can only be expressed in a few ways (e.g. \textit{overdraft}, \textit{direct\_debit}, \textit{swimming\_pool}), while others can have much more varied surface semantic realisations, (e.g. \textit{make, not\_working}). Table~\ref{tab:intents-ontology} in Appendix~\ref{sec:appendix1} provides an estimation of the semantic variability of each intent (module).

\begin{table*}[!t]
\footnotesize
\def\arraystretch{0.99}
\centering
\begin{tabularx}{\linewidth}{l X r}
\toprule
\rowcolor{Gray}
\textbf{Example} & \textbf{Intents}  & \textbf{Slots (\textit{Values)}} \\
\cmidrule(lr){2-3} \cmidrule(lr){2-2} 
\textit{How much less did I spend on Amazon} &how\_much, less, &date\_period (\textit{current year}),\\
\textit{during the current year?} &transfer\_payment &company\_name (\textit{Amazon}) \\
 \vspace{-2mm}
 & & \\
\textit{Show me all the transactions from} &request\_info &date\_from (\textit{Sunday}), \\
\textit{Sunday to Monday please} &transfer\_payment &date\_to (\textit{Monday}) \\
 \vspace{-2mm}
 & & \\
\textit{Hi there, what I want is setting up a 50£} &greet, make, &amount\_of\_money (\textit{50£}), company\_name (\textit{Eon}),\\
\textit{direct debit with Eon for the next 2 months} &direct\_debit &date\_period (\textit{next 2 months}) \\
 \vspace{-2mm}
 & & \\
\textit{Can I make a reservation for 4 adults in} &make, &adults (\textit{4}), rooms (\textit{2}),\\
\textit{2 rooms, from the 1st of June to the 7th?} &booking &date\_from (\textit{1st of June}), date\_to (\textit{7th}) \\
\bottomrule
\end{tabularx}
%\vspace{-1.5mm}
\caption{\nlup examples combining several intents and slots.}
\label{tab:intent-slot-examples} 
%\vspace{-2mm}
\end{table*}

\begin{table*}[t]
\centering
\def\arraystretch{0.93}
\small{
\begin{tabularx}{\linewidth}{l ccccccY}
\toprule
\rowcolor{Gray}
{} & \textbf{Number of}  & \textbf{Number of} & \textbf{Number of} & \textbf{Avg. intents} & \textbf{Avg. slots} & \textbf{Type-token} & \textbf{Semantic} \\
\rowcolor{Gray}
\textbf{Dataset} & \textbf{examples}  & \textbf{intents} & \textbf{slots} & \textbf{per example} & \textbf{per example} & \textbf{ratio (TTR)} & \textbf{diversity} \\
\cmidrule(lr){2-8}
 \textsc{atis} & {5,871} & {18} & {47} & {1$\pm$0.08} & {3.3$\pm$1.61} & {0.043} & {0.202} \\
 \textsc{snips} & {14,484} & {7} & {39} & {1} & {2.6$\pm$1.05} & {0.154} &{0.336} \\
 \textsc{oos} & {23,700} & {151} & {0} & {1} & {0} & {0.148} & {0.254} \\
 \textsc{banking77} & {13,083} & {77} & {0} & {1} & {0} & {0.125} & {0.209} \\
 \nlup & {3,080} & {62} & {17} & {2.01$\pm$1.25} & {0.65$\pm$0.95} & {0.268} & {0.367} \\
\bottomrule
\end{tabularx}
}
%\vspace{-1.5mm}
\caption{Comparison of \nlup with other popular NLU datasets; \textsc{atis} \cite{Hemphill:1990}, \textsc{snips} \cite{Coucke:18}, \textsc{oos} \cite{Larson:2019emnlp} and \textsc{banking77} \cite{Casanueva:2020ws}}.
\label{tab:dataset-comparison}
%\vspace{-2.5mm}
\end{table*}

\subsection{Slots}
\label{sec:slots}

\nlup further includes a rich set of 17 slots, defined in Table~\ref{tab:slots-ontology} in Appendix~\ref{sec:appendix1}. Table~\ref{tab:intent-slot-examples} displays several \nlup examples where complex combinations of intents and slots occur, showcasing how \nlup might provide a much more challenging environment for the evaluation of joint ID and SL models in future research. 

Following the design of previous standard SL datasets \cite{Hemphill:1990, Coucke:18, CoopeFarghly2020}, we provide \textit{span annotations for slots}. On top of of this, to also support training and evaluation of SL models which are not span-based, we also provide \textit{value annotations} (or \textit{canonical values} as named by \citet{rastogi2019towards}) for times, dates, and numeric values.

Similarly to intent modules, slots can also be divided into the \textit{generic} ones (e.g. \textit{time}, \textit{date}) and the \textit{domain-specific} ones (e.g \textit{company\_name}, \textit{rooms}, \textit{kids}), see Table~\ref{tab:slots-ontology}. Again, this distinction allows for the cross-domain reusability of annotated data.

\subsection{Data Collection and Annotation}

Previous NLU datasets have usually relied on \textit{crowdworkers}, aiming to collect a large number of examples, and typically optimising for quantity over quality. However, even with much simpler ontologies, workers are prone to make annotation mistakes, leading to very noisy datasets \cite{Eric:2019arxiv}. In addition, when workers are asked to rephrase a sentence, they often change its semantic meaning or tend to provide rephrasings with extremely low lexical variability \cite{kang-etal-2018-data}.

\nlup reflects true production requirements and focuses on data quality. Instead of relying on {crowdworkers}, 4 highly skilled annotators with dialogue and NLP expertise, also familiar with production environments, collected, annotated, and corrected the data. The process started by defining the ontology for \banking and \hotels. Then, real user examples were fully anonymised and reannotated following the defined ontology. Finally, new examples were created in order to cover less frequent intents and slots, aiming at creating realistic and semantically varied sentences with new combinations of intents and slots. 

%%This yields in a dataset with a very high lexical and semantic diversity.

\subsection{Comparison with Other NLU Datasets}

Aiming to reflect the differences between \nlup and the most popular ToD NLU datasets, Table~\ref{tab:dataset-comparison} compares their general statistics. Since the focus of \nlup is on curated high-quality data, \nlup covers a fewer number of examples than the other datasets, but it is evident that \nlup is the only real multi-intent dataset: it averages {2.01} intents per example with a high standard deviation. In addition, \nlup is the only dataset that combines a large set of intents with a large set of slots.

In order to asses the quality and diversity of the NLU data, we include two additional metrics: 1) Type-Token Ratio (TTR) \cite{jurafsky} which measures lexical diversity) and \textit{semantic diversity}. %NOTE: i need a better way to call this but is a clustering metric i came up with so i dont have a name
Both metrics are computed for the set of examples sharing an intent, weighted by the frequency of that intent\footnote{Note that \textsc{atis} has some intents with a single example: for these intents the TTR score would be 1. Weighting by the intent frequency avoids these intents dominating the metric.} and finally averaged over intents. The semantic diversity per intent is computed as follows: (i) sentence encodings, obtained by the \textit{ConveRT} sentence encoder \cite{Henderson:2020convert},\footnote{See Appendix~\ref{app:sen_enc} for a short description of ConveRT.} are computed for the set of sentences sharing the same intent; (ii) the centroid  of these encodings is then computed; (iii) finally, the average cosine distance from each encoding to the centroid is computed. The overall scores clearly indicate that \nlup offers a much higher lexical and semantic diversity than previous datasets, which should also render it more challenging for current SotA NLU models.\footnote{\textsc{snips} also shows high semantic diversity, but this is mostly due to the high frequency of named entities.}

%due to having been collected by domain experts taking diversity into account.

\section{Experiments and Results}
\label{sec:exp}
In hope to establish \nlup as a more challenging production-oriented testbed for dialogue NLU, especially in low-data scenarios, we evaluate a series of current cutting-edge models for both NLU tasks: intent detection (\S\ref{sec:id_setup}) and slot labeling (\S\ref{sec:sl_setup}). Our aim is to assess and analyse their performance across different setups, and provide solid baseline reference points for future evaluations on \nlup.

\rparagraph{Data Setups}
Unless noted otherwise, for both tasks we adopt the standard $K$-fold cross-validation as done e.g. by \citet{Liu:2019iwsds}. Through such \textit{folding} evaluation, (i) we avoid overfitting to any particular test set and (ii) we ensure more stable results with smaller training and test data (i.e., when simulating low-data regimes typically met in production) through averaging over different folds.\footnote{Due to folding, variations in results with different random seeds were negligible, even in lowest-data setups.}

The experiments are run with $K=20$ (\textbf{20-Fold}) and $K=10$ (\textbf{10-Fold}), where we train on 1 fold and evalute on the remaining $K-1$ folds. These setups simulate different degrees of data scarcity: e.g., the average training fold comprises $\approx100$ examples for \banking and $\approx50$ for \hotels for 20-Fold experiments, and twice as much for 10-Fold experiments. Besides these \textit{low-data training setups}, we also run experiments in a \textbf{Large}-data setup, where we train the models on merged $9$ folds, and evaluate on the single held-out fold.\footnote{Effectively, Large-data experiments can be seen as 10-Fold experiments with swapped training and test data.} The key questions we aim to answer with these data setups are: Which NLU models are better adapted to low-data scenarios? How much does NLU performance improve with the increase of annotated NLU data? How challenging is \nlup in low-data versus large-data scenarios?

\rparagraph{Domain Setups}
Further, experiments are run in the following domain setups: (i) \textit{single-domain} experiments where we only use the \banking or the \hotels portion of the entire dataset; (ii) \textit{both-domain} experiments (termed \textsc{all}) where we use the entire dataset and combine the two domain ontologies (see Table~\ref{tab:dataset-stats}); (iii) \textit{cross-domain} experiments where we train on the examples associated with one domain and test on the examples from the other domain, keeping only \textit{shared} intents and slots for evaluation. The key questions we aim to answer are: Are there major performance differences between the two domains and can they be merged into a single (and more complex) domain? Is it possible to use examples labeled with generic intents from one domain to boost another domain, effectively increasing reusability of data annotations and reducing data scarcity?

$F_1$ (micro) is the main evaluation measure in all ID and SL experiments.

\subsection{Intent Detection: Experimental Setup}
\label{sec:id_setup}
We evaluate two groups of SotA intent detection models: (i) \textit{MLP-Based}, and (ii) \textit{QA-Based} ones. 

\rparagraph{MLP-Based ID Baselines}
%% A standard transfer learning paradigm \cite{Ruder:2019transfer} fine-tunes a pretrained language model such as BERT \cite{Devlin:2018arxiv} or RoBERTa \cite{Liu:2019arxiv} on the annotated task data. 
\newcite{Casanueva:2020ws} and \newcite{Gerz:2021arxiv} have recently shown that, for the ID task, full and expensive fine-tuning of large pretrained models such as BERT \cite{Devlin:2018arxiv} or RoBERTa \cite{Liu:2019arxiv} is not needed to reach strong ID performance. As an alternative, they propose a much more efficient \textit{MLP-based} approach to intent detection which works on par or even outperforms full fine-tuning on the ID task.\footnote{Our preliminary results on the \nlup dataset corroborated these findings from prior work; due to a large number of experiments, we thus opt for this more efficient yet also very effective approach to ID.} In a nutshell, the idea is to use fixed/frozen ``off-the-shelf'' universal sentence encoders such as ConveRT \cite{Henderson:2020convert} or Sentence-BERT \cite{Reimers:2019emnlp} models to encode input sentences. A standard multi-layer perceptron (MLP) classifier is then learnt on top of the sentence encodings. 

Two core differences to the previous work stem from the fact that we now deal with the multi-label ID task: \textbf{1)} to this end, we replace the output \textit{softmax} layer with the \textit{sigmoid} layer; and \textbf{2)} we define a threshold $\theta$ which determines the final classification: only intents with probability scores $\geq\theta$ are taken as positives. This way, the hyper-parameter $\theta$ effectively controls the trade-off between precision and recall of the multi-label classifier.

We comparatively evaluate several widely used state-of-the-art (SotA) sentence encoders, but remind the reader that this decoupling of the MLP classification layers from the fixed encoder allows for a much wider empirical comparison of sentence encoders in future work. The evalauted sentence encoders are: \textbf{1)} \textbf{\textsc{ConveRT}} \cite{Henderson:2020convert}, which produces 1,024-dimensional sentence encodings; \textbf{2)} \textbf{\textsc{LaBSE}} \cite{Feng:2020labse} (768-dim); \textbf{3)} \textbf{\textsc{RobL-1B}} (1,024-dim) and \textbf{4)} \textbf{\textsc{LM12-1B}} (384-dim) \cite{Reimers:2019emnlp,Thakur:2021naacl}. For completeness, we provide brief descriptions of each encoder in our evaluation, along with their public URLs, in Appendix~\ref{app:sen_enc}, and refer the reader to the original work for more details about each sentence encoder.

\rparagraph{QA-Based ID Baselines}
Another group of SotA ID baselines reformulates the ID task into the (extractive) question-answering (QA) problem \cite{qanlu,Fuisz:2022arxiv}. This QA-oriented reformatting then allows for additional specialised QA-tuning of large PLMs. In a nutshell, the idea is to (i) fine-tune the original PLM such as BERT/RoBERTa on readily available large general-purpose QA data such as SQuAD \cite{Rajpurkar:16}, and then (ii) further fine-tune this general QA model with in-domain ID data. This strategy has recently shown very strong performance on single-label ATIS data \cite{qanlu}. 

The main `trick' is to reformat the input ID examples into the following format: \textit{``yes. no. [SENTENCE]''} and pose a question such as: \textit{``is the intent to ask about [INTENT]?''} (see Appendix~\ref{sec:appendix1} for the actual questions associated with each intent, also shared with the dataset). Here, \textit{[SENTENCE]} is the placeholder for the actual input sentence, and \textit{[INTENT]} is the placeholder for a short manually defined text (akin to language modeling prompts \cite{Liu:2021survey}, see again Appendix \ref{sec:appendix1}) which briefly describes the intent. The QA formulation lends itself naturally to the multi-label ID setup as each `intent-related' question is posed separately. In other words, for each input example and for each of the $L$ intents in the ontology the QA model must extract \textit{yes} or \textit{no} as the answer, where correct intent labels are the ones for which the answer is \textit{yes}.\footnote{For instance, for the input sentence \textit{``I need to increase my overdraft''} from the \banking domain, we would pose all 48 questions associated with each of the $L=48$ intents in \banking, where the QA model should extract \textit{yes} as the answer for intents \textit{change}, \textit{overdraft} and \textit{more\_higher\_after}, and extract \textit{no} for the remaining 45 intents in \banking.}  We note that our work is the first to apply and evaluate the QA approach on multi-label ID.

We experiment with two pretrained language models, both fine-tuned on the SQuAD2.0 dataset \cite{rajpurkar2018know} before additional QA-tuning on \nlup examples converted to the aforementioned QA format: \textbf{\textsc{RobB-QA}} uses RoBERTa-Base as the underlying LM, while \textbf{\textsc{Alb-QA}} relies on the more compact ALBERT \cite{Lan:2019albert}.

%% (IV, return to CR)
%%\footnote{Again, we note that these underlying models can be substituted by other QA-tuned LMs.} 

%% The three main results table
\begin{table*}[!t]
\def\arraystretch{0.96}
\centering
{\scriptsize
\begin{tabularx}{\linewidth}{l YYY YYY YYY}
\toprule
  {} & \multicolumn{3}{c}{\bf \textsc{banking}} & \multicolumn{3}{c}{\bf \textsc{hotels}} & \multicolumn{3}{c}{\bf \textsc{all}} \\
  \cmidrule(lr){2-4} \cmidrule(lr){5-7} \cmidrule(lr){8-10}
\textbf{Setup}$\rightarrow$ & \textbf{20-Fold}  & \textbf{10-Fold} & \textbf{Large} & \textbf{20-Fold}  & \textbf{10-Fold} & \textbf{Large} & \textbf{20-Fold}  & \textbf{10-Fold} & \textbf{Large} \\
\cmidrule(lr){2-10}
 %\cmidrule(lr){2-4} \cmidrule(lr){5-7} \cmidrule(lr){8-10}
 %\cmidrule(lr){1-10}
 \rowcolor{Gray}
 {\textbf{Sentence Encoder}$\downarrow$} & \multicolumn{9}{c}{\bf MLP-Based Baselines} \\
 %\cmidrule(lr){2-4} \cmidrule(lr){5-7} \cmidrule(lr){8-10}
  \cmidrule(lr){2-10}
 \textsc{ConveRT} & {58.6} & \underline{70.2} & \underline{90.3} & {52.3} & {63.1} & \underline{82.8} & \underline{58.6} & \underline{70.2} & \underline{88.9} \\
 \textsc{LaBSE}* & {54.8} & {66.6} & {88.7} & {48.9} & {58.9} & {82.3} & {55.4} & {66.1} & {87.0} \\
 \textsc{RobL-1B}* & {56.8} & {68.4} & {87.4} & \underline{55.2} & \underline{64.2} & {81.8} & {57.3} & {67.7} & {86.2} \\
 \textsc{LM12-1B}* & \underline{59.1} & {69.0} & {87.8} & {53.5} & {62.8} & {79.5} & {58.4} & {68.2} & {86.0} \\
 \cmidrule(lr){2-10}
 \rowcolor{Gray}
 {\textbf{QA-Pretrained Model}$\downarrow$} & \multicolumn{9}{c}{\bf QA-Based Baselines} \\
\cmidrule(lr){2-10}
\textsc{RobB-QA}* & {\bf 80.3} & {\bf 85.6} & {\bf 93.1} & {\bf 67.4} & {\bf 73.3} & {\bf 86.7} & {\bf 79.5} & {\bf 84} & {\bf 91.8} \\
\textsc{AlbB-QA}* & {76.6} & {82.1} & {92.0} & {60.7} & {67.2} & {85.1} & {75.5} & {80.8} & {90.6} \\
\bottomrule
\end{tabularx}
}%
%\vspace{-1.5mm}
\caption{$F_1$ scores ($\times$100\%) of benchmarked state-of-the-art intent detection models on NLU++ in three data setups (see \S\ref{sec:id_setup}). We also refer to \S\ref{sec:exp} for the brief descriptions of each sentence encoder (for MLP-based baselines) and the two QA-pretrained models. *All models were retrieved from the HuggingFace model repository \cite{Wolf:2019hf}, with exact model URLs available in Appendix~\S\ref{app:sen_enc} and Appendix~\S\ref{app:idqa}. The overall best-performing model per column is in \textbf{bold}, while the best-performing MLP-based model per column is \underline{underlined}.}
\label{tab:main_ic}
%\vspace{-1.5mm}
\end{table*}

\begin{table}[!t]
\def\arraystretch{0.96}
\centering
{\scriptsize
\begin{tabularx}{\linewidth}{l Y Y Y}
\toprule
\rowcolor{Gray}
 {\textbf{\textsc{ConVEx}}} & \textbf{20-Fold}  & \textbf{10-Fold} & \textbf{Large} \\
\cmidrule(lr){1-4}
 %\cmidrule(lr){2-4} \cmidrule(lr){5-7} \cmidrule(lr){8-10}
 %\cmidrule(lr){1-10}
 %\cmidrule(lr){2-4} \cmidrule(lr){5-7} \cmidrule(lr){8-10}
  %\cmidrule(lr){1-1}
 \banking & {30.1} & {40.0} & {68.1} \\
 \hotels & {29.7} & {40.0} & {64.5} \\
 \all & {34.0} & {45.2} & {71.4} \\
 \cmidrule(lr){1-4}
\rowcolor{Gray}
 {\textbf{QA-Based}: \textsc{RobB-QA}} & \textbf{20-Fold}  & \textbf{10-Fold} & \textbf{Large} \\
\cmidrule(lr){1-4}
\banking & {50.5} & {56.7} & {70.2} \\
 \hotels & {48.1} & {52.4} & {70.4} \\
 \all & {55.5} & {53.6} & {72.1} \\
\bottomrule
\end{tabularx}
}%
%\vspace{-1.5mm}
\caption{$F_1$ scores ($\times$100\%) on the \nlup SL task for \textsc{ConVEx} \cite{Henderson:2021convex} and a \textit{QA-Based} approach \cite{qanlu} across different domains and data setups.} 
\label{tab:main_ve}
%\vspace{-1.5mm}
\end{table}

\begin{table}[!t]
\def\arraystretch{0.97}
\centering
{\scriptsize
\begin{tabularx}{\linewidth}{l Y Y}
\toprule
%\rowcolor{Gray}
 & \textbf{\banking$\rightarrow$\hotels}  & \textbf{\hotels$\rightarrow$\banking}\\
\cmidrule(lr){2-2} \cmidrule(lr){3-3}
\rowcolor{Gray}
\textbf{MLP-Based} & {} & {} \\
\textsc{ConveRT} & {75.4} & {65.2} \\
\textsc{LM12-1B} & {67.3} & {49.2} \\
\hdashline
\rowcolor{Gray}
\textbf{QA-Based} & {} & {} \\
\textsc{Alb-QA} & {76.7} & {72.7} \\
\textsc{RobB-QA} & {\bf 79.3} & {\bf 74.2} \\
 %\cmidrule(lr){2-4} \cmidrule(lr){5-7} \cmidrule(lr){8-10}
 %\cmidrule(lr){1-10}
 %\cmidrule(lr){2-4} \cmidrule(lr){5-7} \cmidrule(lr){8-10}
  %\cmidrule(lr){1-1}
\bottomrule
\end{tabularx}
}%
%\vspace{-1.5mm}
\caption{$F_1$ scores of \textit{cross-domain} intent detection experiments, evaluating performance on the set of 26 intents shared by the two domains. \textit{Large}-data setup.} 
\label{tab:id_cd}
%\vspace{-1.5mm}
\end{table}

\rparagraph{ID: Training and Evaluation} All MLP-based baselines rely on the same training protocol and hyper-parameters in all data and domain setups. The MLP classifier consists of 1 hidden layer of size 512, and is trained via binary cross-entropy loss for 500 epochs with the batch size of 32 and the dropout rate is 0.6. We use the standard AdamW optimizer \cite{Loschilov:2018iclr} with the learning rate of 0.003 and linear decay; weight decay is 0.02. The threshold $\theta$ is set to 0.4.\footnote{These hyper-parameters were selected based on preliminary experiments with a single (most efficient) sentence encoder \textsc{lm12-1B} and training only on Fold 0 of the 10-Fold \banking setup; they were then propagated without change to all other MLP-based experiments with other encoders and in other setups. We repeated the similar hyper-parameter search procedure for QA-based models, using \textsc{Alb-QA}.\label{fn:repeat}.}

For QA models, we largely follow \newcite{qanlu} and fine-tune all models for 5 epochs, using AdamW; the learning rate of $2e\!-\!5$ with linear decay; weight decay is 0; batch size is 32.

%$F_1$ (micro) is the main ID evaluation measure.

\subsection{Slot Labeling: Experimental Setup}
\label{sec:sl_setup}

For slot labeling, we benchmark two current SotA models: (i) \textit{ConvEx} \cite{Henderson:2021convex}, as a SotA span-extraction SL model and (ii) the QA-based SL model \cite{qanlu} based on \textsc{RobB-QA}, which operates similarly to QA-based ID baselines discussed in \S\ref{sec:id_setup}, and relies on the same fine-tuning regime as our QA-based ID baselines. Again, we refer the reader to the original work for further details, and provide brief descriptions in Appendix~\ref{app:sl_models}.

\subsection{Results and Discussion}
\label{sec:discussion}
Main results with all the evaluated baselines are summarised in Table~\ref{tab:main_ic} (for ID) and Table~\ref{tab:main_ve} (SL). 

\rparagraph{ID: MLP versus QA Models}
First, the comparisons among only MLP-based models reveal that \textbf{1)} all sentence encoders offer ID performance in similar, reasonably narrow score intervals (e.g., the variations in $F_1$ scores between all sentence encoders are typically below 4-6 $F_1$ points in all setups), and \textbf{2)} that \textsc{ConveRT} is the best-performing sentence encoder on average, which corroborates findings from prior work on other ID datasets \cite{Casanueva:2020ws,Wu:2020salesforce}.

One very apparent and important indication in the reported results is the superiority of QA-based ID models over their MLP-based competitors. QA-based models largely outperform MLP-Based baselines in all domain setups, as well as in all data setups. The gains are visible even in Large-data setups, but the benefits of QA-based ID are immense in the lowest-data 20-Fold setups: e.g., 12 $F_1$ points over the strongest MLP ID model on \hotels and 20 $F_1$ points on \banking. 

Moreover, the use of larger underlying LMs might push the scores with QA even further: using SQuAD-tuned Roberta-Large (\textsc{RobL-QA}) instead of Base (\textsc{RobB}) yields further gains -- e.g., $F_1$ rises from 85.6 to 87.8 on 10-Fold \banking, and similar trends are observed in other low-data setups. 

\rparagraph{Slot Labeling}
In the SL task, the QA-based model also demonstrates its superiority, again with huge gains in low-data 20-Fold and 10-Fold setups, confirming that such QA-based or prompt-based methods \cite{Liu:2021survey,Gao:2021prompts} are especially well suited for low-data setups. The use of manually defined questions/prompts, which are typically easy to write by humans, combined with the expressive power of QA-based task formatting yields immense gains on low-resource dialogue NLU.

Given these very promising ID and SL results on \nlup, our work also calls for further and more intensive future research on QA-based models for dialogue NLU. However, we note that QA-based ID and SL methods do come with efficiency detriments, especially with larger intent and slot sets: the model must copy the input utterance and run a separate answer extraction for each intent/slot from the set, which is by several order of magnitudes more costly at both training and inference than MLP-based models. A promising future research avenue is thus to investigate combined approaches that could combine and trade off the performance benefits of QA-based models and the efficiency advantages of, e.g., MLP-based ID.

\rparagraph{Low-Data vs. Large-Data}
We also note that scores on both tasks, as reported in Tables~\ref{tab:main_ic}-\ref{tab:main_ve}, leave ample room for improvement in NLU methodology in future work, especially on SL (even in Large-data setups), and in low-data setups.

%While it could be argued that QA-based models make use of extra resources by using the questions defined in \ref{tab:slots-ontology}, which is specially crucial in few-shot scenarios, the gains obtained by this approach are immense.

\rparagraph{Cross-Domain Experiments} We also verify potential reusability of annotated data across domains with a simple ID experiment, where we train ID models on \banking and evaluate on \hotels, and vice versa. The results are summarised in Table~\ref{tab:id_cd}. Besides (again) indicating that QA-based models outscore MLP-based ID, the results also suggest that for some \textit{generic} intents it is possible to meet high ID performance without any in-domain annotations. For instance, we observe particularly high scores for highly generic and reusable intent modules such as \textit{change}, \textit{how}, \textit{how\_much}, \textit{thank}, \textit{when}, and \textit{affirm}, all with per-intent $F_1$ scores of $\geq 90$. We hope that these preliminary results might inspire similar ontology (re)designs in future work.

%In order to overcome the smaller size of the dataset, we evaluate it in a cross-validation setup as in \cite{Liu:2019iwsds}. 

%%We do the experiments in a very low data settings (coz is what we want in these sort of domains) and we do cross validation eval (to avoid overfitting to the test set)

%EXPERIMENTS TODO: 

%IC and VE baselines on our data, focused on low data regimes

%Cross domain experiments (train in one domain and test on the other)

%% The main IC table with all the results

\section{Conclusion}
We have presented \nlup, a novel dataset for task-oriented dialogue (ToD) NLU that overcomes the shortcomings of previous NLU evaluation sets. \nlup presents a multi-intent and slot-rich ontology, defines generic and domain-specific intents and slots to promote data reusability, and it focuses on the creation of high-quality complex examples and annotations collected by dialogue experts. Experimental results show that \nlup raises the bar with respect to current NLU benchmarks, helping better discriminate and compare the performance of current state-of-the-art NLU models, particularly in low-data setups. We hope that \nlup will be valuable in guiding future modeling efforts for ToD NLU, both in academia and in industry.

%We believe that this can help to improve current models for both academia and in industry.

\rparagraph{Limitations and Future Work}
This work has shown that a better design of the intent set can improve data reusability. However, the current ontology does not cover generic sets of intents exhaustively, and we acknowledge a (sometimes) fine line between truly generic intents versus intents `anecdotally' shared by two domains (e.g., \textit{refund}). Further, the boundaries of some generic intents can sometimes be unclear and difficult to annotate, even for expert annotators.\footnote{For example, boundaries for intents like \textit{greet}, \textit{why} and \textit{change} are clear, while others such as \textit{make} or \textit{not\_working} are more prone to ambiguity and different interpretation.} Future work should try to ground the set of generic intents.

Further, we believe that span-based annotation might be sub-optimal for canonical values such as \textit{times} and \textit{dates}, where small differences in the span would lead to evaluation errors but would not suppose a problem for the value to be parsed. In addition, separating \textit{time} and \textit{date} intervals in different slots increases the difficulty of the annotations and models need to learn a more conflicting set of slots. Further, \nlup currently provides fine-grained slots such as \textit{date\_from}, \textit{date\_to} and \textit{date} to enable more complex scenarios, but such a design might slow down annotation process and make it cumbersome. Future work includes rethinking the SL task for these slots.

Finally, while single-turn NLU is more data-efficient and easier to model, some user utterances only make sense in the presence of context from the previous system utterance. While some previous datasets \cite{CoopeFarghly2020} deal with this issue with the help of extra annotations indicating if a slot has been requested, in this work we opt for using \textit{non-contextualised} slots such as \textit{number} and \textit{time} and let the policy handle the contextualisation. However, future work should start looking into NLU datasets composed by \textit{system + user} turns.

\section*{Acknowledgements}
We are grateful to our colleagues in PolyAI for many fruitful discussions and suggestions. We also thank the anonymous reviewers for their helpful feedback and comments on the presentation.


\section*{Ethical Considerations}
\textcolor{black}{PolyAI Limited} is ISO27k-certified and fully GDPR-compliant.

\textit{Before data collection:} all the data has been collected by workers of PolyAI Limited and all the annotators are also employees of PolyAI Limited. %authors of this paper

\textit{During data collection:} we did not include any personal information (e.g. personal names or addresses) and all the examples that included any had been fully anonymised or removed from the dataset. All the names in the dataset are created by randomly concatenating names and surnames from the list of the top 10K names from the US registry. Upon collection, the dataset has undergone an additional check by the internal Ethics committee of the company. NLU++ is licensed under CC-BY-4.0.

%\newpage
\bibliography{anthology,custom}
\bibliographystyle{acl_natbib}

\clearpage
\appendix
\section{Appendix: Ontology}
The complete ontology of \nlup is provided in Table~\ref{tab:intents-ontology} and Table~\ref{tab:slots-ontology}.
\label{sec:appendix1}
\begin{table*}[t]\centering
\scriptsize
\begin{tabular}{l|lccll}
& & &LEXICAL &\\
INTENT &DESCRIPTION-QUESTION &DOMAIN &DIVERSITY &CATEGORY \\
\hline\hline
affirm &is the intent to affirm something? &general &medium &General dialogue \\
deny &is the intent to deny something? &general &medium &acts  \\
dont\_know &is the intent to say I don't know? &general &high & \\
acknowledge &is the intent to acknowledge what was said? &general &medium & \\
greet &is the intent to greet someone? &general &high & \\
end\_call &is the intent to end call or say goodbye? &general &high & \\
handoff &is the intent to speak to a human or hand off? &general &high & \\
thank &is the intent to thank someone? &general &medium & \\
repeat &is the intent asking to repeat the previous sentence? &general &medium & \\
\hline
cancel\_close\_leave &is the intent asking about canceling or closing something? &general &high &Actions \\
change &is the intent to change or modify something? &general &high & \\
make &is the intent to make, open, apply, set up or activate something? &general &high & \\
\hline
request\_info &is the intent to ask or request some information? &general &high &Questions \\
how &is the intent asking how to do something? &general &medium & \\
why &is the intent to ask why something happened or needs to be done? &general &medium & \\
when &is the intent to ask about when or what time something happens? &general &medium & \\
how\_much &is the intent asking about some quantity or how much? &general &medium & \\
how\_long &is the intent asking about how long something takes? &general &medium & \\
\hline
not\_working &is the intent asking about something wrong, missing or not working? &general &high &General adjectives \\
lost\_stolen &is the intent asking about something being lost or stolen? &general &medium & \\
more\_higher\_after &is the intent to indicate something more, higher, after or increasing? &general &medium & \\
less\_lower\_before &is the intent to indicate something less, lower, before or decreasing? &general &medium & \\
new &is the intent asking about something new? &general &medium & \\
existing &is the intent asking about something that already exists? &general &medium & \\
limits &is the intent asking about some sort of limit? &general &medium & \\
\hline
savings &is the intent asking about the savings account? &banking &low &Domain specific\\
current &is the intent asking about the current account? &banking &low &adjectives  \\
business &is the intent to ask something about the business account? &banking &low & \\
credit &is the intent asking about something related to credit? &banking &low & \\
debit &is the intent asking about something related to debit? &banking &low & \\
contactless &is the intent to ask about contactless? &banking &low & \\
international &is the intent to ask about something related to international issues? &banking &medium & \\
\hline
account &is the intent asking about some account? &banking &low &Domain specific\\
transfer\_payment &is the intent to ask about something related to a transfer,&banking &low &nouns/entities \\
&payment or deposit? & & & \\
appointment &is the intent to ask about something about an appointment? &banking &medium & \\
arrival &is the intent to ask about the arrival of something? &banking &medium & \\
balance &is the intent to ask about balance? &banking &medium & \\
card &is the intent to ask about something related to a card or cards? &banking &low & \\
cheque &is the intent to ask about cheque? &banking &low & \\
direct\_debit &is the intent to ask about direct debit? &banking &low & \\
standing\_order &is the intent asking about a standing order? &banking &low & \\
fees\_interests &is the intent to ask about fees or interests? &banking &medium & \\
loan &is the intent to ask about loans? &banking &low & \\
mortgage &is the intent asking about mortgage? &banking &low & \\
overdraft &is the intent to ask about ovedraft? &banking &low & \\
withdrawal &is the intent to ask about withdrawals? &banking &low & \\
pin &is the intent to ask something about the pin number? &banking &low & \\
refund &is the intent to ask about some refund? &banking, hotels &low & \\
check\_in &is the intent to ask about check in? &hotels &medium & \\
check\_out &is the intent to ask about check out? &hotels &medium & \\
restaurant &is the intent to ask something related to restaurant? &hotels &medium & \\
swimming\_pool &is the intent to ask something related to the swimming pool? &hotels &low & \\
parking &is the intent to ask something related to parking? &hotels &low & \\
pets &is the intent to ask something related to pets? &hotels &medium & \\
accesibility &is the intent to ask something related to accessibility? &hotels &medium & \\
booking &is the intent to talk about some booking? &hotels &medium & \\
wifi &is the intent to ask something related to wifi or wireless? &hotels &low & \\
gym &is the intent to ask something related to gym? &hotels &low & \\
spa &is the intent to ask something related to spa or beauty services? &hotels &high & \\
room\_ammenities &is the intent to ask something related to some room amenities? &hotels &high & \\
housekeeping &is the intent to talk about housekeeping issues? &hotels &medium & \\
room\_service &is the intent to talk about room service? &hotels &medium & \\
\bottomrule
\end{tabular}
\caption{Intents ontology}\label{tab:intents-ontology}
\end{table*}

\begin{table*}[t]\centering
\scriptsize
\begin{tabular}{l|lcc}\toprule
SLOT &DESCRIPTION-QUESTION &DOMAIN \\
\hline\hline
date &What is the specific date mentioned in this sentence? &general \\
date\_period &What is the time period in days, months or years mentioned in this sentence? &general \\
date\_from &What is the start date of some period mentioned in this sentence? &general \\
date\_to &What is the end date of some period mentioned in this sentence? &general \\
time &What is the specific time in the day mentioned in this sentence? &general \\
time\_from &What is the start time of some time period mentioned in this sentence? &general \\
time\_to &What is the end time of some time period mentioned in this sentence? &general \\
time\_period &What is the time period in hours or minutes mentioned in this sentence? &general \\
person\_name &What is the name of a person mentioned in this sentence? &general \\
number &What is the number without context mentioned in this sentence? &general \\
\hline
amount\_of\_money &What is the specific amount of money mentioned in this sentence? &banking \\
company\_name &What is the name of some sort of company mentioned in this sentence? &banking \\
shopping\_category &What is the category of some expense mentioned in this sentence? &banking \\
\hline
kids &what is the number of kids mentioned in this sentence? &hotels \\
adults &what is the number of adults mentioned in this sentence? &hotels \\
people &What is the number of people mentioned in this sentence? &hotels \\
rooms &What is the number of rooms mentioned in this sentence? &hotels \\
\bottomrule
\end{tabular}
\caption{Slots ontology}\label{tab:slots-ontology}
\end{table*}

\section{Appendix: Sentence Encoders in Intent Detection Experiments}
\label{app:sen_enc}
\noindent \textbf{\textsc{ConveRT}} \cite{Henderson:2020convert} is trained with the conversational response selection objective \cite{Henderson:2019acl} on large Reddit data \cite{AlRfou:2016arxiv,Henderson:2019arxiv}, spanning more than 700M \textit{(context, response)} sentence pairs. Thanks to its naturally conversational pretraining objective, it has been shown to be especially well-suited for conversational tasks such as intent detection \cite{Casanueva:2020ws} and slot labelling \cite{CoopeFarghly2020}. It outputs 1,024-dim sentence encodings.

\vspace{0.5mm}
\noindent - \url{github.com/davidalami/ConveRT}

\vspace{1.2mm}
\noindent \textbf{\textsc{LaBSE}.} Language-agnostic BERT Sentence Embedding (LaBSE) \cite{Feng:2020labse} adapts pretrained multilingual BERT (mBERT) \cite{Devlin:2018arxiv} using a dual-encoder framework \cite{yang2019ijcai} with larger embedding capacity (i.e., a shared multilingual vocabulary of 500k subwords). While LaBSE is the current state-of-the-art multilingual encoder, it also displays very strong monolingual English performance \cite{Feng:2020labse}. It produces $768$-dim sentence encodings.

\vspace{0.5mm}
\noindent - \url{huggingface.co/sentence-transformers/LaBSE}

\vspace{1.2mm}
\noindent \textbf{\textsc{RobL-1B}} and \textbf{\textsc{LM12-1B}} \cite{Reimers:2019emnlp,Thakur:2021naacl} are sentence encoders which fine-tune the pretrained Roberta-Large (\textsc{RobL}) language model \cite{Liu:2019arxiv} and the 12-layer MiniLM \cite{wang2020minilm}, respectively, again using a contrastive dual-encoder framework \cite{Reimers:2019emnlp}. The models are fine-tuned on a set of more than 1B sentence pairs: this set comprises various data such as Reddit 2015-2018 comments \cite{Henderson:2019arxiv}, Natural Questions \cite{nq}, PAQ (question, answer) pairs \cite{paq}, to name only a few.\footnote{In a nutshell, the contrastive fine-tuning task which combines all the heterogeneous datasets is as follows: given a `query' sentence from each sentence pair, and a set of $R$ randomly sampled negatives plus $1$ true positive (the sentence from the same pair), the model should predict which sentence from the set of $R+1$ sentences is actually paired with the query sentence in the dataset. The full list of all datasets along with the exact model specifications is at: \\ \url{huggingface.co/sentence-transformers/all-roberta-large-v1}.} \textsc{RobL-1B} outputs 1,024-dim encodings, while \textsc{LM12-1B} produces 384-dim encodings. 

We opted for those two models in particular as one represents a class of large sentence encoders (\textsc{RobL-1B}), and the other is lightweight (\textsc{LM12-1B}), while both display very strong performance in a myriad of sentence similarity and semantic search tasks, see \url{www.sbert.net/docs/pretrained_models.html}.

\vspace{0.5mm}
\noindent - \url{huggingface.co/sentence-transformers/all-roberta-large-v1} \\
\noindent - \url{huggingface.co/sentence-transformers/all-MiniLM-L12-v1}

\section{Appendix: QA-Pretrained Models}
\label{app:idqa}
We rely on the same SQuAD-tuned language models as \newcite{qanlu}.  \textbf{\textsc{RobB-QA}} can be found online at: \url{https://huggingface.co/deepset/roberta-base-squad2}; \textbf{\textsc{alb-QA}} is available at: \url{https://huggingface.co/twmkn9/albert-base-v2-squad2}

\section{Appendix: Slot Labeling Baselines}
\label{app:sl_models}
\noindent \textbf{\textsc{ConVEx}} \cite{Henderson:2021convex} demonstrates strong SL performance, especially in few-shot settings. It is pretrained on a pairwise cloze task extracted from the Reddit examples \cite{Henderson:2019arxiv}, and the majority of the pretrained model's parameters in \textsc{ConVEx} are kept frozen during fine-tuning, making it an extremely efficient model. We adopt the suggested hyper-parameters from \newcite{Henderson:2021convex}.

\vspace{1.2mm}
\noindent \textbf{QA-Based:} \citet{qanlu} train an extractive QA-based model to extract the spans of the slots from the input user utterance as answers to manually defined natural language questions (one per slot). It follows the same idea as QA-based ID models. We also provide such questions for each slot along with \nlup for model training and inference: see the questions in Table~\ref{tab:slots-ontology}. 

%We use RoBERTa-base \cite{Liu:2019roberta} as PLM and we do the first stage of QA training with SQuAD \cite{Rajpurkar:16}.

\end{document}